\documentclass[letterpaper, 10pt, conference]{ieeeconf}
\IEEEoverridecommandlockouts    %
\usepackage{graphicx}
\usepackage{tabularx}
\usepackage{caption}
\usepackage{subcaption}
\usepackage{makecell} %
\usepackage{xcolor} %
\usepackage{amsmath} %
\usepackage{amsfonts, bm} %
\usepackage{cuted} %

\usepackage{eso-pic}
\newcommand{\mycopyrighttext}{%
  \footnotesize
  \noindent
  \textcopyright~2025 IEEE. Personal use of this material is permitted.
  Permission from IEEE must be obtained for all other uses, in any current
  or future media, including reprinting/republishing this material for
  advertising or promotional purposes, creating new collective works,
  for resale or redistribution to servers or lists, or reuse of any
  copyrighted component of this work in other works.\\
  International Conference on Advanced Robotics and Mechatronics (ICARM), 1-3 August, 2025.
}

\AtBeginDocument{%
  \AddToShipoutPicture*{%
    \AtPageUpperLeft{%
      \put(55, -40){%
        \parbox{\dimexpr\textwidth-4pt\relax}{\raggedright \mycopyrighttext}
      }%
    }
  }
}

\title{\LARGE \bf
LAURON VI: A Six-Legged Robot for Dynamic Walking
}

\author{Christian Eichmann$^{1}$, Sabine Bellmann$^{1}$, Nicolas Hügel$^{1}$, Louis-Elias Enslin$^{1}$,\\ Carsten Plasberg$^{1}$, Georg Heppner$^{1}$, Arne Roennau$^{1,2}$ and Ruediger Dillmann$^{1}$%
\thanks{$^{1}$FZI~Research Center for Information Technology,
Haid-und-Neu-Str.\ 10–14, 76131~Karlsruhe, Germany.
{\tt\small eichmann@fzi.de}.\newline
$^{2}$Machine Intelligence and Robotics Lab at the Karlsruhe Institute for Technology (KIT),
76131~Karlsruhe, Germany.
{\tt\small roennau@kit.edu}.\newline
The research leading to these results has received funding from the DLR Space Administration under grant agreement No.\ 50RA1730, 50RA2026, and 50RA2404 by the German Bundestag.
}%
}

\begin{document}

\maketitle
\thispagestyle{empty}
\pagestyle{empty}

\begin{abstract}
    Legged locomotion enables robotic systems to traverse extremely challenging terrains.
    In many real-world scenarios, the terrain is not that difficult and these mixed terrain types introduce the need for flexible use of different walking strategies to achieve mission goals in a fast, reliable, and energy-efficient way.
    Six-legged robots have a high degree of flexibility and inherent stability that aids them in traversing even some of the most difficult terrains, such as collapsed buildings.
    However, their lack of fast walking gaits for easier surfaces is one reason why they are not commonly applied in these scenarios.

    This work presents LAURON VI, a six-legged robot platform for research on dynamic walking gaits as well as on autonomy for complex field missions.
    The robot's 18 series elastic joint actuators offer high-frequency interfaces for Cartesian impedance and pure torque control.
    We have designed, implemented, and compared three control approaches: kinematic-based, model-predictive, and reinforcement-learned controllers.
    The robot hardware and the different control approaches were extensively tested in a lab environment as well as on a Mars analog mission.
    The introduction of fast locomotion strategies for LAURON VI makes six-legged robots vastly more suitable for a wide range of real-world applications.

\end{abstract}


\section{INTRODUCTION} \label{sec:introduction}

Inspired by nature, legged locomotion is of great value for robotics, given its robustness and adaptability.
Stable locomotion is especially advantageous in difficult terrain.
We have seen great leaps in four-legged robotic locomotion in recent years.
Improvements in actuator technology, as well as control algorithms, allow for successfully solving a wide range of tasks with legged robots.
They are successfully applied in the fields of exploration, search-and-rescue, transportation and (industrial) monitoring.

\begin{figure}[t]
    \includegraphics[width=\columnwidth]{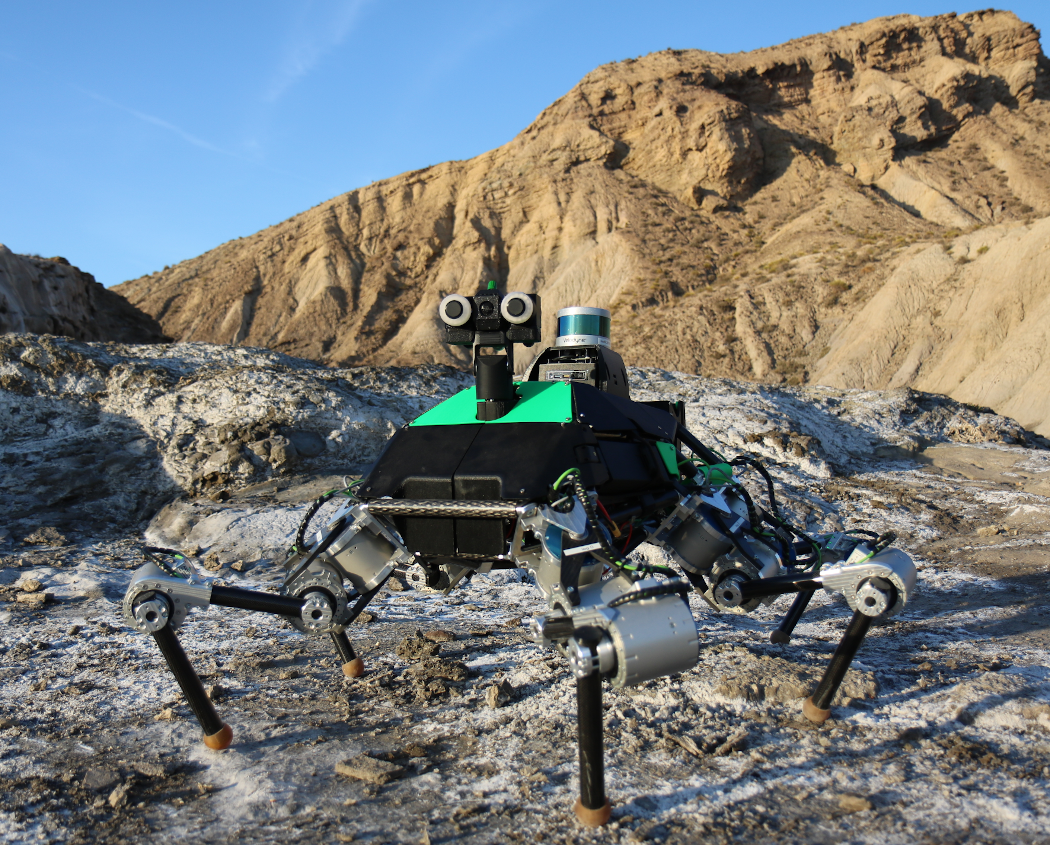}
    \caption{LAURON VI in the Tabernas Desert, Spain}
    \label{fig:Spain}
\end{figure}

Given their high stability, hexapods enjoy continuous interest in the scientific community for a diverse set of applications and with different control strategies \cite{coelho_trends_2021}.
LAURON~V is a 26 degree of freedom (DoF) hexapod controlled by a behavior-based control approach \cite{roennau_lauron_2014}.
Capable of switching between an insectoid and mammalian leg configuration, the Bullet robot can utilize a variety of statically stable walking patterns \cite{tam_openshc_2020}.
The Bruce robot is built using linear series elastic actuators and utilizes a central pattern generator, which can produce a 0.5 $ms^{-1}$ top speed on flat terrain~\cite{steindl_bruce_2020}.
HAntR, with its four DoF in each leg, is focused on statically stable locomotion through rough terrain \cite{cizek_design_2021}.
It can carry impressive payloads weighing up to 85 \% of its own body mass.
Besides walking, hexapods can be used in other challenging use cases like climbing.
Lin  et~al.\ use a multi-layered optimization approach to climb chimney-like structures with their six-legged robot using friction while avoiding obstacles \cite{lin_optimization_2019}.
Research regarding learned walking behaviors on hexapods is mainly focused on employing reinforcement learning trained in simulation \cite{schilling_decentralized_2020,azayev_blind_2020}  and using spiking neural networks \cite{lele_learning_2020,naya_spiking_2021}.

While there is some interest in the scientific community in the locomotion of six-legged robots, their main application is as robust sensor platforms for other research.
For example, Dupeyroux  et~al.\ use a light-weight hexapod with a regular tripod walking pattern to research and recreate the navigational capabilities of desert ants, using only a minimal sensor setup \cite{dupeyroux_antbot_2019}.
Agheli  et~al.\ utilize the great stability and range of motion for maintenance tasks \cite{agheli_shero_2014}.

With the great advancements in four-legged locomotion, the field of quadrupeds may serve as inspiration for the development of hexapods.
The commercially available four-legged robot ANYmal \cite{hutter_anymal_2016} is already used for a wide range of applications, both in research \cite{arm_scientific_2023} and industry \cite{ishigami_anymal_2021}.
Researchers have implemented different control strategies for locomotion on the platform, like optimization-based controllers \cite{xin_robust_2021} and learned behaviors \cite{rudin_learning_2021,jenelten_dtc_2024}.
Both approaches are also applicable to the wheeled variant of the robot \cite{bjelonic_whole-body_2021,lee_control_2023}.
Focused on forward speed, the MIT Cheetah III developed by Bledt  et~al.\ uses a whole-body motion controller to achieve a cost of transport of 0.45 \cite{bledt_mit_2018}.
Di Carlo  et~al.\ use model-predictive control based on Single Rigid-Body Dynamics for feet force optimization to achieve 3~m/s forward speeds and impressive lateral and angular movements with the same robot \cite{di_carlo_dynamic_2018}.
With Mini Cheetah \cite{katz_mini_2019}, Katz  et~al.\ presented a highly dynamic 9~kg quadruped with the ability to do backflips and various dynamic gaits~\cite{kim_highly_2019}.

These new insights and approaches regarding dynamic legged locomotion are rarely applied to hexapods.
One reason for this is the lack of platforms capable of dynamic motions.

In this paper, we present LAURON~VI, a modern six-legged walking robot the size of commercially available quadrupeds.
The LAURON VI robot is built utilizing 18 series elastic actuators.
This paper presents different control strategies and showcase the capabilities of the robotic system in simulation as well as on the hardware:
A simple trajectory-based controller to execute different walking gaits, a model-predictive controller optimizing ground reaction forces, and a walking policy trained using reinforcement learning.

Section \ref{sec:hardware} describes the robotic hardware design, Section \ref{sec:walking} focuses on the approach used to generate the walking gaits.
In Section \ref{sec:results}, the experimental evaluation and results are presented.
Section \ref{sec:conclusion} concludes with a discussion of the presented work and provides an outlook for future research.

\section{Design of a Dynamic Hexapod} \label{sec:hardware}

\begin{table}[!t]
  \renewcommand{\arraystretch}{1.3}
  \vspace{0.15cm}
  \caption{LAURON VI — Specifications}
  \label{table:specs}
  \centering
  \begin{tabular}{|l|l|}
  \hline
  \multicolumn{2}{|l|}{LAURON VI} \\
  \hline
  Type & Six-Legged Walking Robot\\
  \hline
  No. of Joints per Leg & 3 (alpha, beta, gamma)\\
  \hline
  Main actuators & ANYdrives: series elastic actuator from \\
  &  ANYbotics, 35 Nm (alpha, gamma) or \\
  & 70 Nm (beta) peak torque\\
  \hline
  No. of Head Joints & 2 (pan, tilt)\\
  \hline
  Size [footprint] & 130 cm x 120 cm \\
  \hline
  Height & 68 cm (head), 35 cm (hip) \\
  \hline
  Weight & 42 kg excluding batteries \\
  \hline
  Max Payload & 12 kg\\
  \hline
  Power Supply & two or four 22.2 V 16Ah LiPo batteries\\
  & or 48 V external power supply \\
  \hline
  Power Consumption & Standing: 250 W, Walking 400 W \\
  \hline
  On-board PC & Intel Core i9-9900K CPU, 8×3.6 GHz, \\
  & 32 GB RAM \\
  \hline
  Additional Sensors & LORD microstrain 3DM-GX5-25 IMU, \\
  & optional Velodyne HDL-32E \\
  \hline
  \end{tabular}

  \vspace{-5pt}

\end{table}

LAURON VI is a six-legged, electrically actuated walking robot and the newest member of our LAURON family of bionic hexapods.
It uses ROS 2 \cite{macenski_robot_2022} as its middleware and an EtherCAT bus for real-time communication. %
Table~\ref{table:specs} summarizes the most important technical specifications.

\subsection{Requirements} \label{sec:requirements}

To achieve fast and dynamic walking, the drives need to be able to create and sustain high torques.
The leg workspaces need to be wide enough to allow for long strides and to perform other tasks such as manipulation or foot point planning in difficult terrain with limited available footholds.
The robot should have a high impact resistance and be easy to maintain, improving longevity and viability for demanding missions.
Additionally, it should be able to carry payloads like sensors, computing, power, and more.
The main body should encapsulate the components for computing, power, networking, and sensors while being light\-weight and robust.
The robot should provide the ability to mount exchangeable payloads on its back to increase the range of tasks it can perform.

The six legs ought to be similar in construction and easy to exchange to improve the maintainability of the robot and decrease complexity.
Insectoid kinematics shall allow a wide stance and stable walking in diverse terrain.
Therefore, a reasonably sized workspace needs to be guaranteed for each leg.

LAURON VI should be powered directly via an external power supply or by LiPo batteries.
Providing four battery bays allows for increased runtime and adds the ability to hot-swap batteries without shutting down the robot,
as only two batteries are needed simultaneously to power the robot.

\begin{figure}[b]
    \vspace{-0.3cm}
    \centering
    \begin{subfigure}[b]{0.32\columnwidth}
        \centering
        \includegraphics[width=\textwidth]{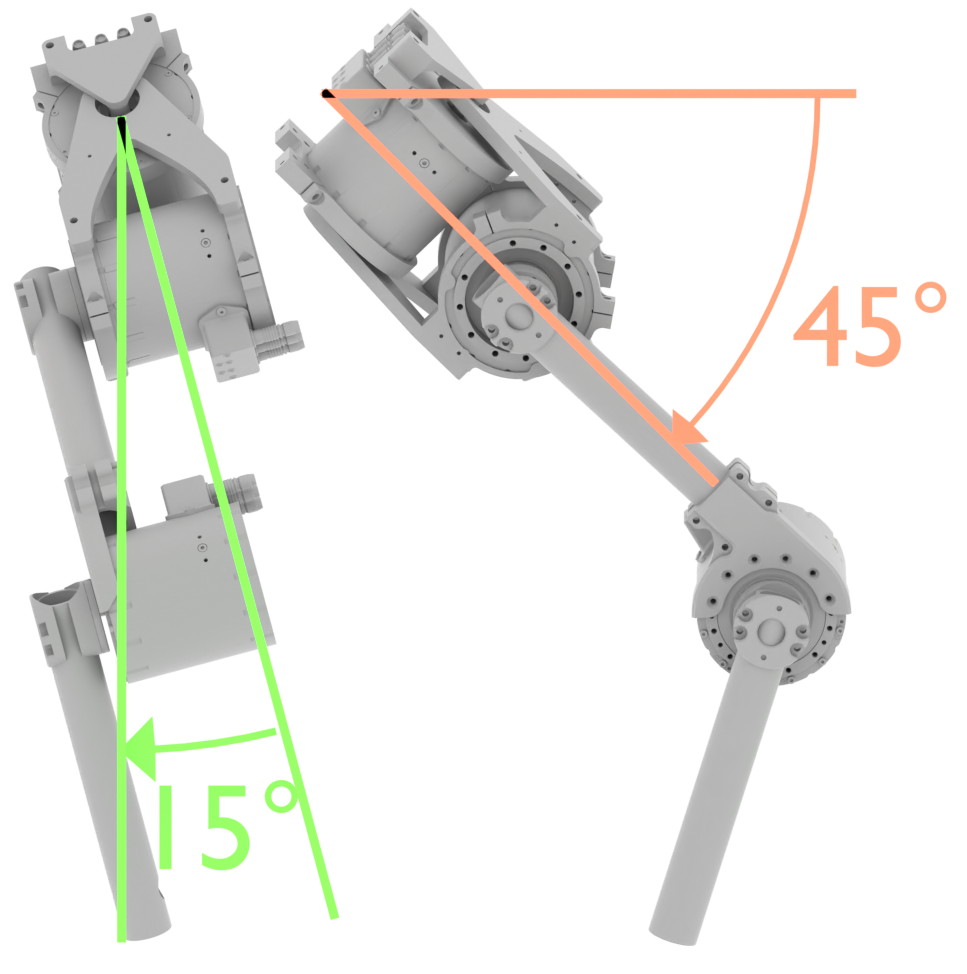}
    \end{subfigure}
    \begin{subfigure}[b]{0.32\columnwidth}
        \centering
        \includegraphics[width=\textwidth]{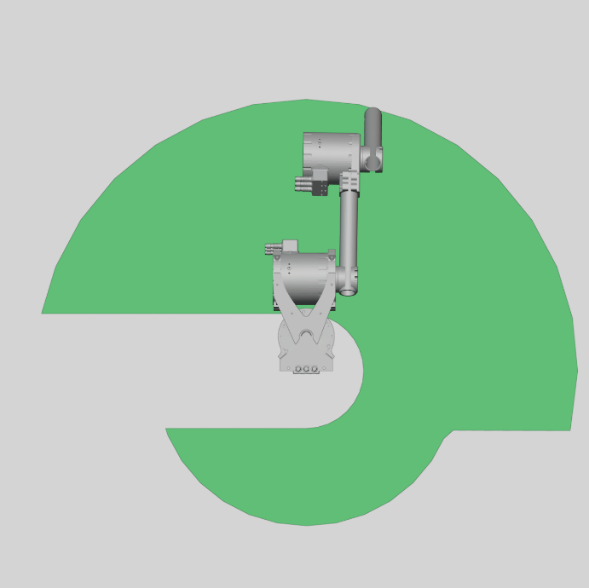}
    \end{subfigure}
    \begin{subfigure}[b]{0.32\columnwidth}
        \centering
        \includegraphics[width=\textwidth]{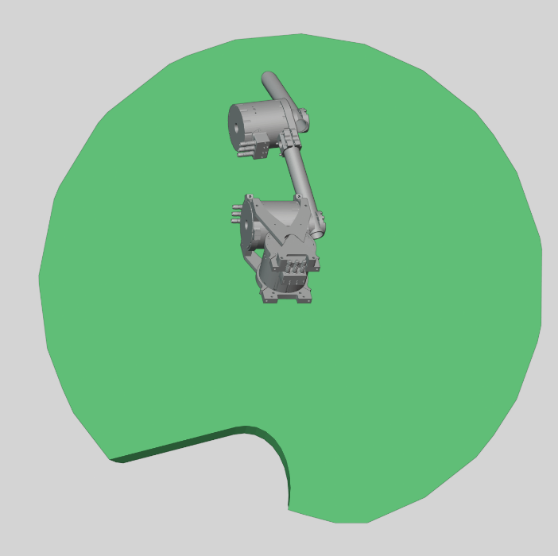}
    \end{subfigure}
    \caption{Visualization of the leg's tilt (left) and a comparison of the straight (middle) and tilted (right) leg's workspace at a stance height of 0.3~m}
    \label{fig:workspace}
\end{figure}

\begin{figure}[h]
    \vspace{0.3cm}
    \centering
    \begin{subfigure}[b]{0.49\columnwidth}
        \centering
        \includegraphics[width=\textwidth]{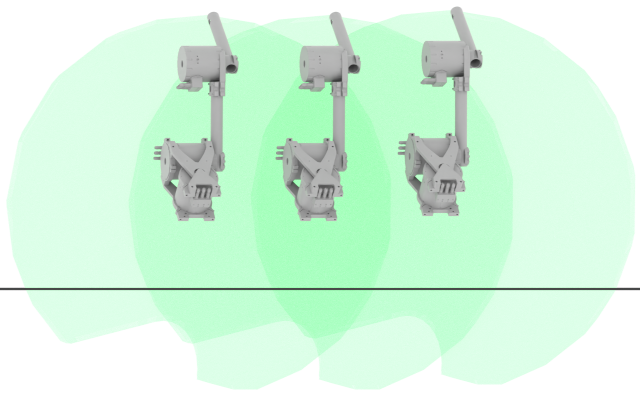}
    \end{subfigure}
    \begin{subfigure}[b]{0.49\columnwidth}
        \centering
        \includegraphics[width=\textwidth]{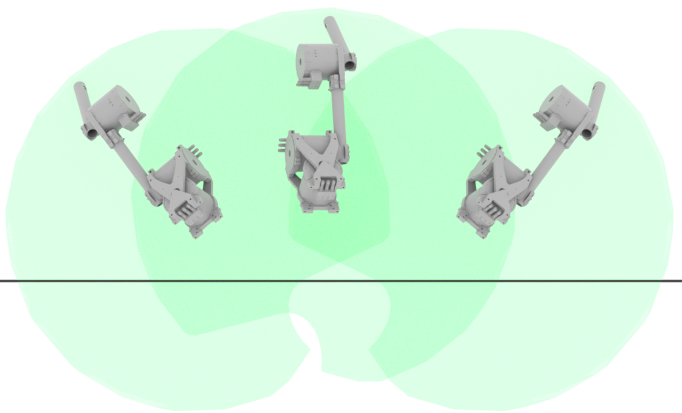}
    \end{subfigure}
    \caption{Combined workspace comparison for parallel (left) and skewed (right) leg attachments at a stance height of 0.3~m}
    \label{fig:workspaces}
\end{figure}

\subsection{Leg Placement} \label{sec:leg}
The legs need to be designed and mounted onto the robot body to maximize their workspace while keeping expected torques within reasonable limits to conserve energy.
There are two different designs, one left and one right leg design.
Each leg is actuated by three ANYdrive \cite{hutter_anymal_2016} series elastic actuators.
Only four screws and two cables connect each leg to the main body, making replacement quick and easy.

Sampling the reachable volume for given leg attachments allows to judge a configuration's viability.
In this case, configuration means the roll and pitch angles at the leg attachment points and the stance height.
A roll angle of 60~degrees combined with a pitch of -15~degrees yield the biggest continuous reachable area at around 0.35~m stance height.
However, given its visible deviation from what is considered an insectoid stance, the slightly sub-optimal roll angle of 45~degrees is chosen instead, to preserve the robots insectoid characteristics.
This configuration, as well as its workspace, is shown in Fig.~\ref{fig:workspace}.

Attaching the legs parallel to each other onto a robot body results in a lot of overlap between the workspaces.
Fanning them out by adding a yaw of $\pm$40~degrees to the front and hind legs increases the covered area.
The area can be further increased by mirroring the hind legs.
The combined improvements are visualized in Fig.~\ref{fig:workspaces}.

\subsection{Topology Optimization} \label{sec:topology}
\begin{figure}[t]
    \centering
    \begin{subfigure}[b]{\columnwidth}
        \centering
        \includegraphics[width=\textwidth]{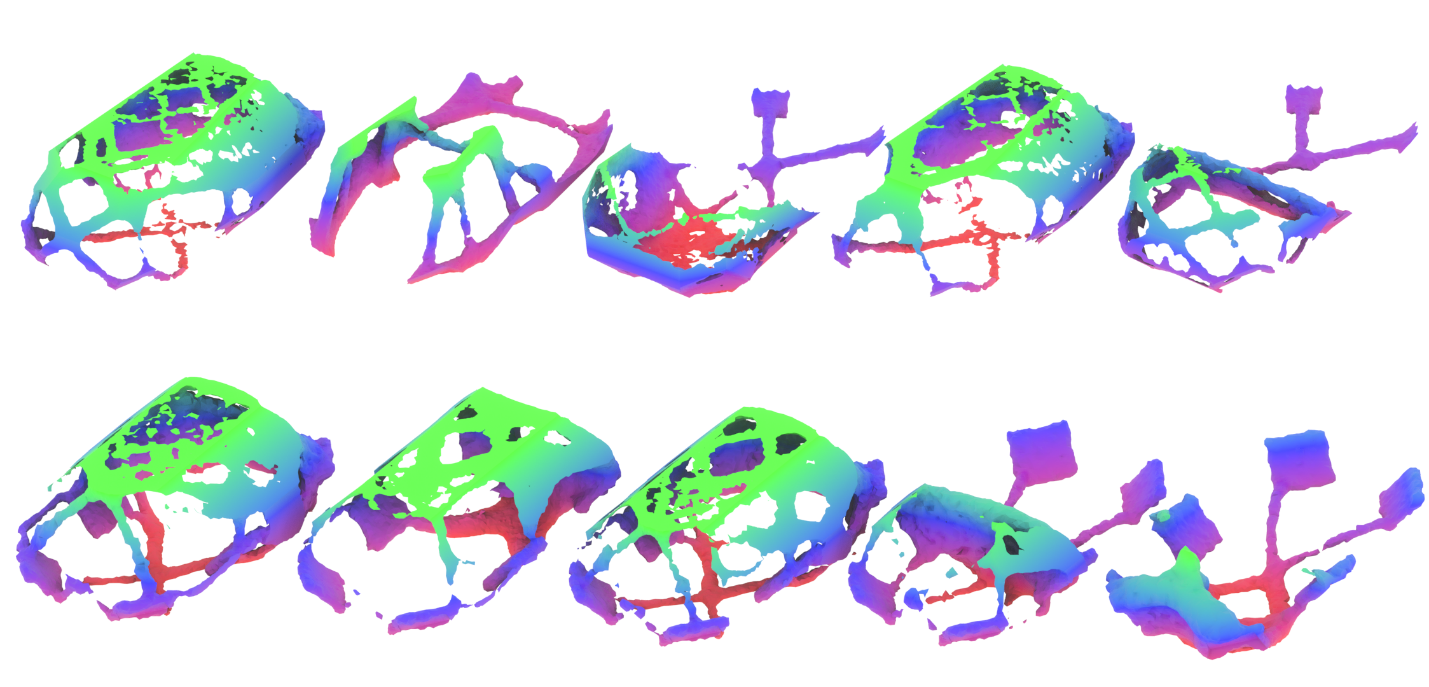}
    \end{subfigure}
    \begin{subfigure}[b]{0.6\columnwidth}
        \centering
        \includegraphics[width=\textwidth]{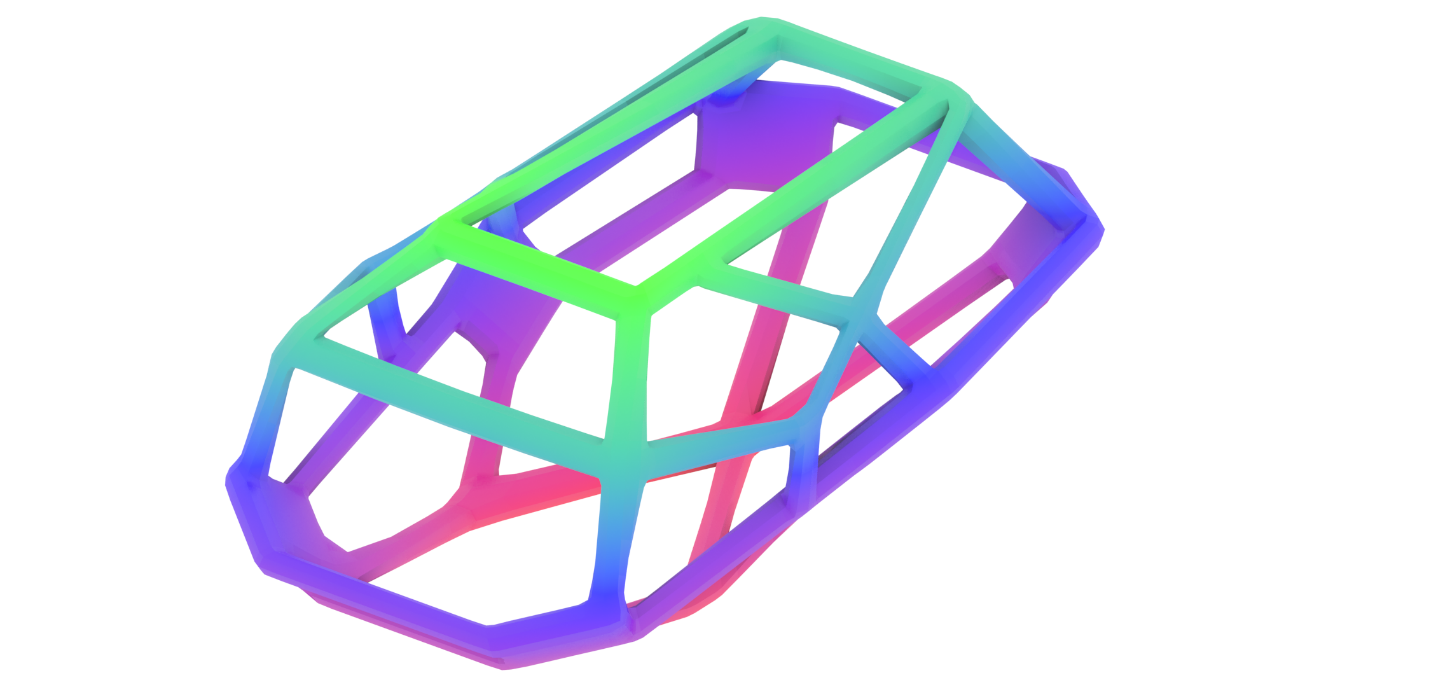}
    \end{subfigure}
    \caption{Topology optimization results for different forces acting on the robot body, with the final body structure at the bottom}
    \label{fig:topology}
\end{figure}

The main body structure of LAURON~VI is the result of a topology optimization, taking into consideration the expected forces induced by walking, external forces like carrying the robot by hand, and the required space to house internal components.
Using a Gazebo simulation \cite{koenig_design_2004} with the ODE physics engine \cite{smith_open_2001}, expected walking forces and moments acting on the robot's body were calculated.
A number of topology optimizations for different observed force profiles acting on the leg mount points were generated using  Autodesk Inventor~\cite{noauthor_autodesk_nodate}, resulting in a wide range of specialized body structures.
Combining these structures delivers an optimized robot body able to withstand at least twice the forces generated in simulation.
Fig.~\ref{fig:topology} shows a selection of 10 optimization results as well as the structure of the combined robot body.

Given that only pressure and tension rods are needed, the body could be designed with a truss structure comprised of carbon fiber tubes.
Custom-made parts milled from aluminum and 3d printed out of PETG connect the pipes, profiles, and legs.
In a second iteration, aluminum profiles were incorporated into its back to serve as attachment points for different payloads.

\begin{figure}[bh!]
    \includegraphics[width=\columnwidth]{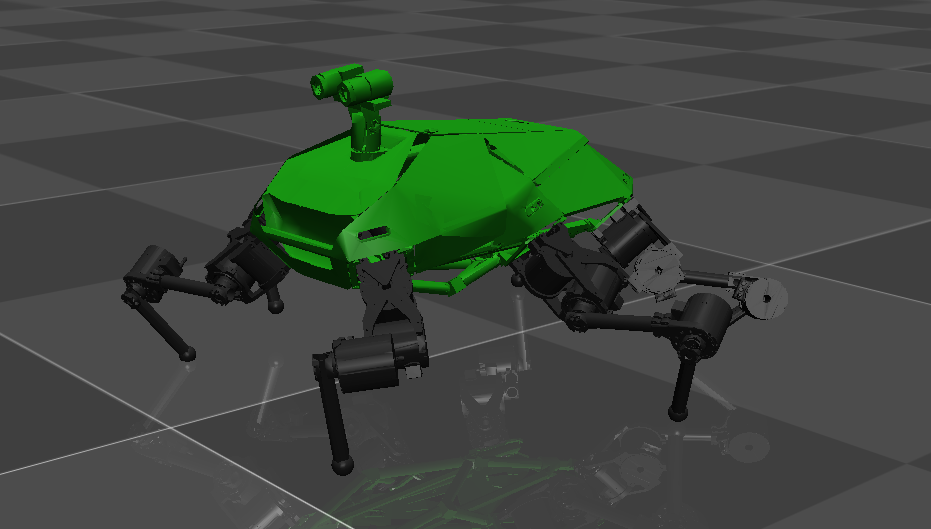}
    \caption{LAURON VI walking in the MuJoCo simulation using a reinforcement learned policy.}
    \label{fig:mujoco}
\end{figure}

\begin{figure*}[!t]
  \normalsize
  \vspace{0.2cm}
  \begin{equation}
    \label{eq:model}
      \left[
        \begin{array}{c}
          \dot{\boldsymbol{\Theta}}\\
          \dot{\mathbf{p}} \\
          \dot{\boldsymbol{\omega}} \\
          \ddot{\mathbf{p}}\\
          \dot{g}
        \end{array}
      \right]%
      =%
      \left[
        \begin{array}{ccccc}
          \mathbf{0}_3 & \mathbf{0}_3 & \mathbf{R}_z(\psi) & \mathbf{0}_3 & \vec{\mathbf{0}}_3^{\top} \\
          \mathbf{0}_3 & \mathbf{0}_3 & \mathbf{0}_3 & \mathbf{1}_3 & \vec{\mathbf{0}}_3^{\top} \\
          \mathbf{0}_3 & \mathbf{0}_3 & \mathbf{0}_3 & \mathbf{0}_3 & \vec{\mathbf{0}}_3^{\top} \\
          \mathbf{0}_3 & \mathbf{0}_3 & \mathbf{0}_3 & \mathbf{0}_3& \left[0\ \ 0\ \ 1\right]^{\top} \\
          \vec{\mathbf{0}}_3 & \vec{\mathbf{0}}_3 & \vec{\mathbf{0}}_3 & \vec{\mathbf{0}}_3 & 0
        \end{array}
      \right]%
      \left[
        \begin{array}{c}
          \boldsymbol{\Theta} \\
          \mathbf{p} \\
          \boldsymbol{\omega} \\
          \dot{\mathbf{p}}\\
          g
        \end{array}
      \right]%
      +%
      \left[
        \begin{array}{ccc}
          \mathbf{0}_3 & \cdots &\mathbf{0}_3 \\
          \mathbf{0}_3 & \cdots &\mathbf{0}_3 \\
          \mathbf{I}^{-1}\left[\mathbf{r}_0\right]_{\times} & \cdots & \mathbf{I}^{-1}\left[\mathbf{r}_5\right]_{\times} \\
          \mathbf{1}_3 / m_r & \cdots & \mathbf{1}_3 / m_r  \\
          \vec{\mathbf{0}}_3 & \cdots & \vec{\mathbf{0}}_3
        \end{array}
      \right]%
      \left[
        \begin{array}{c}
          \mathbf{f}_{grf,0} \\
          \mathbf{f}_{grf,1} \\
          \mathbf{f}_{grf,2} \\
          \mathbf{f}_{grf,3} \\
          \mathbf{f}_{grf,4} \\
          \mathbf{f}_{grf,5}
        \end{array}
      \right]\\
  \end{equation}
\end{figure*}

\section{Walking Controllers} \label{sec:walking}

From the previous LAURON robots, we have learned that position-based kinematic control of six-legged robots provides a lot of stability but lacks in movement speed.
Experience shows that mobile exploration in real-world scenarios requires both stability and speed to solve realistic tasks in mixed environments.
Behavior-based control was sufficient for traversing unknown terrain without the need for intensive planning, but its high number of parameters makes usage and optimization difficult \cite{kerscher_behaviour-based_2008}.

Modern hexapods have to be able to walk fast and dynamic on flat terrain and utilize more stable gaits for rough and challenging surfaces.
For fast walking, the forward movement is more important than sideways motion or rotations, based on the assumption that it is mainly used to quickly traverse big, flat stretches of fairly even terrain.
Therefore, optimizations improving forward motions can be made, even if the performance for sideways or rotational movements is impacted negatively.
For more challenging terrain, a wide stance and a quasi-static gait can be employed.
Given that the hexapod's main strength is the high stability in challenging terrain, this work aims to improve their utilization by ensuring points of interest can be reached quickly, even if vast stretches of flat terrain need to be traversed first.

With LAURON VI, we strive to build a platform to research dynamic walking for six-legged robots.
Given that both optimal control and reinforcement learning approaches show promising results with four-legged robots and
to showcase the flexibility of the robot, three different control strategies were implemented and compared:
\begin{itemize}
    \item A kinematic-based controller moving the robot through adaptable foot point trajectories
    \item A model-predictive controller based on Single Rigid-Body Dynamics (SRBD) and
    \item A policy learned by reinforcement learning.
\end{itemize}
Fig.~5 shows a simulation based on MuJoCo \cite{todorov_mujoco_2012}, which is used to speed up gait design, software development, and testing.
The simulation is designed to provide identical interfaces and sensor readings as the hardware drivers.

The reuse of existing components is one major focus of the software architecture, as components can be used for the different controllers.
All controllers interact with the actuators by commanding target points and target forces on a per-leg level, as well as motor gains to change the leg's behavior.
The state estimation and processing of sensor data is also shared between the controllers.
The robot is able to do predefined motions to stand up and sit down, has self-collision avoidance and a state machine to prevent dangerous user commands.
\subsection{Simple Kinematic-based Controller} \label{sec:kinematic}
Our kinematic-based controller executes periodic Cartesian swing and stance foot point trajectories to move the robot body.
Different periodic gaits are defined in YAML configurations.
In the swing phase, the legs follow a Bézier curve as a position-based trajectory but are compliant to be able to withstand unexpected shocks.
In stance, the robot moves its feet inverse to the desired body motion and applies an additional feed-forward force to compensate for gravity.
This is achieved by using a Cartesian Impedance Controller for each leg, with adjustable stiffness and damping, and a feed-forward force at the ground contact point.

We apply the derived target forces through joint torques calculated with the Jacobian matrix.
The ANYdrive actuators employ a cascading controller to control joint positions, velocities, and torques.
For all motions, our kinematic-based approach commands a target position derived from the foot trajectory through an analytical solution to the leg's inverse kinematics, a target velocity of zero, and a feet-forward torque as gravity compensation.
The velocity of zero allows the D-gain of the controller to act similar to a damper, while the P-gain approximates a spring constant, allowing an intuitive configuration of the leg's impedance.
\subsection{Model-Predictive Controller} \label{sec:MPC}
To increase speed and energy efficiency compared to the kinematic-based controller, a model-predictive controller (MPC) was developed.
The MPC concept is adopted for six-legged robots from the work on the MIT Cheetah~III by Di~Carlo  et~al.\ \cite{di_carlo_dynamic_2018}.

The controller optimizes the feet forces over the prediction horizon to control the linear and angular velocities, the position, and the orientation of the robot.
The robot state is described with 12 variables: the Euler angles in ZYX representation, where $\psi$ is the yaw, $\theta$ is the pitch, and $\phi$ is the roll, the robot position $\mathbf{p}$, its angular velocities $\bm{\omega}$ and linear velocities $\mathbf{\dot{p}}$.
We use a single rigid-body dynamics model in  the \textit{world-frame} to represent the dynamics.
For both the calculation of the inertia I and the derivative of the Euler angles simplifications are used following \cite{di_carlo_dynamic_2018}.

The state-space form is shown in Equation~\ref{eq:model}, where $\mathbf{f}_{grf,j}$ is the vector of the ground reaction forces (GRF) of leg $j$, which are the opposing forces to the leg forces.

This state space variant of the equation is solely a function of $\psi$ and $\mathbf{r}_{j}$, the latter one being the position of leg $j$:
\begin{equation}
  \label{eq:state_space}
  \dot{\mathbf{x}}(t)
  =
  \mathbf{A}_{c}(\psi)\mathbf{x}(t)
  +
  \mathbf{B}_{c}(\mathbf{r}_{0},\ \ldots,\ \mathbf{r}_{5},\ \psi)\mathbf{u}(t)
\end{equation}
where $\mathbf{A}_{c} \in \mathbb{R}^{13 \times 13}$ is the continuous state space matrix and $\mathbf{B}_{c} \in \mathbb{R}^{13 \times 18}$ is the continuous control matrix.

The minimal and maximal ground reaction forces are set as constraints to limit forces in the stance phases and keep the GRF vector inside a friction pyramid.
GRFs are set to zero in swing phase and $f_{max}$ in stance, with linear interpolation between these values.
The state space and control matrices are discretized using the Zero Order Hold approach.

For the MPC formulation, the state space is substituted into the cost function instead of including it in the constraints, which speeds up the solver \cite{di_carlo_dynamic_2018}:
\begin{equation}
  \label{eq:mpc}
  \begin{aligned}
    \min\limits_{\mathbf{u}}\ \ \ 
    \sum_{i=0}^{k-1}\vert\vert
      \mathbf{A}_{i}\mathbf{x}_{i}
      +
      \mathbf{B}_{i}\mathbf{u}_{i}
      -
      \mathbf{x}_{i+1,\mathbf{des}}
    \vert\vert _{\mathbf{L}_{i}}
    +
    \vert\vert
      \mathbf{u}_{i}
    \vert\vert _{\mathbf{K}_{i}} \\
    \text{subject}\ \text{to}\ 
    \mathbf{f}_{grf,i,min}
    \leq 
    \mathbf{C_i}\mathbf{u}_{i}
    \leq
    \mathbf{f}_{grf,i,max}, i=0\ldots k-1
  \end{aligned}
\end{equation}
where $\mathbf{L_i} \in \mathbb{R}^{13 \times 13}$ is the state weighting matrix and $\mathbf{K_i} \in \mathbb{R}^{18 \times 18}$ the control weighting matrix.
$\mathbf{L_i}$ contains weighting factors for the orientation, position, the angular and linear velocities.
Following the work of Di Carlo  et~al.\ \cite{di_carlo_dynamic_2018}, the optimization problem of Equation~\ref{eq:mpc} is solved by reformulation into a quadratic programming problem, which we solve using the OSQP solver \cite{stellato_osqp_2020}.
The MPC on LAURON~VI runs with a frequency of 100 Hz and has a prediction horizon of 5 steps, which corresponds to 0.5 s.

\begin{figure}[h]
    \vspace{0.2cm}
    \includegraphics[width=\columnwidth]{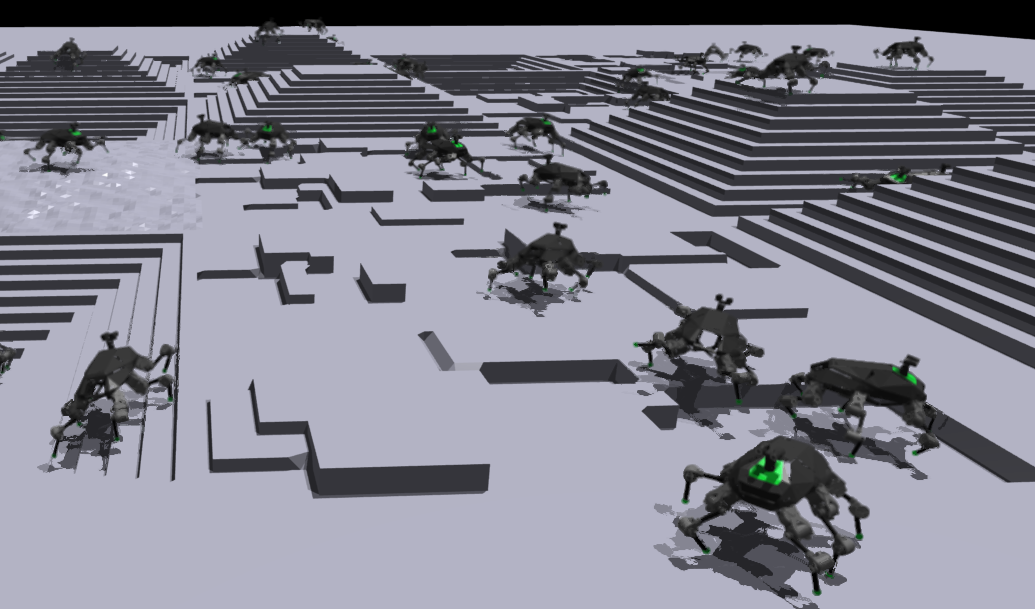}
    \caption{Multiple simulated LAURON VI training using legged\_gym \cite{rudin_legged_gym_2022}.}
    \label{fig:isaac}
\end{figure}

\subsection{Reinforcement Learning} \label{sec:DRL}

Following recent trends, we used reinforcement learning to train a walking policy for LAURON VI.
We build upon the work from Rudin  et~al.\ \cite{rudin_learning_2021}, utilizing their legged\_gym \cite{rudin_legged_gym_2022} code base to create a training curriculum for six-legged locomotion.
LAURON VI was added to the legged\_gym and trained with the same game-like environments but with simpler terrain to compensate for the robot's current lack of depth perception, as shown in Fig.~\ref{fig:isaac}.
This is done to improve the adaption to non-flat terrain while taking the reduced, camera-free sensor setup of LAURON VI into account.
The addition of six depth cameras to the robot and in the curriculum is planned for the future.

\begin{table}[t]
  \renewcommand{\arraystretch}{1.3}
  \vspace{0.15cm}
  \caption{
      Terms of the cost function used for the reinforcement learned controller, many originate from legged\_gym~\cite{rudin_legged_gym_2022} and were adapted for six legs.
      Changed or new terms are marked with *.
    }
  \label{table:costs}
  \centering
  \begin{tabularx}{\columnwidth}{|l X|}
  \hline
  \thead{Cost term} & \thead{Description} \\
  \hline

  \multicolumn{2}{|c|}{Goal tracking rewards} \\
  \hline

  $ e^{-((\Delta {v}_{x})^2 + (\Delta v_{y})^2)  / \sigma} $ & Rewards linear velocity target \\
  \hline

  $ e^{-(\Delta \omega_{z})^2 / \sigma} $ & Rewards angular velocity target \\
  \hline

  \multicolumn{2}{|c|}{Hardware limitation rewards} \\
  \hline

  $ \sum |F_{collision}| $  & Penalizes self-collisions (contacts between different body parts). \\
  \hline

  $ \sum_{i=1}^{18}(\tau^2)$ & * Penalizes executed forces, should reduce energy consumption and heat buildup \\
  \hline

  $ \sum_{i=1}^{18}( ({a}_{t-1} - {a}_{t})^2 ) $ & * Discourages changes in actions to reduce oscillation; $a$ is the action output of the policy \\
  \hline

  $ \sum_{i=1}^{18}( (\omega_{t-1} - \omega_{t} / \Delta t)^2)$ & * Penalizes accelerations in actuators \\
  \hline

$\begin{array} {ll} \sum (|F_{ground}| - F_{limit}) \\  \text{for } |F_{ground}| > F_{limit} \end{array}$ & * Penalizes contact forces above a set limit \\
\hline

  \multicolumn{2}{|c|}{Gait shaping rewards} \\
  \hline

  $ {(v_{z})}^2 $ & Discourages motions in z axis \\
  \hline

  $ (\Delta \omega_{x})^2 + (\Delta \omega_{y})^2 $ & Discourages rotations around x and y axis \\
  \hline

  $ (\overrightarrow{g}_{x,base})^2 + (\overrightarrow{g}_{y,base})^2 $ & Discourages rotations compared to flat ground, $ \overrightarrow{g}_{base} $ is the gravitational vector rotated by the base orientation. \\
  \hline

   $ \sum_{i=1}^{18} (\Delta x) \text{ for } v_{des} < v_{limit} $& * Discourages joint position changes when commanded linear velocity is near zero to reduce movements without commands \\
   \hline

  $|h_{curr} - h_{des}|$ & * Rewards body height close to target, changed from quadratic to absolute error \\
  \hline

  $ \sum_{i=1}^{6} (t_{i,air}) \text{ if new contact} $  & * Rewards based on continuous non-contact time of legs to encourage longer steps, only if commanded velocity is not zero; $t_{i,air}$ is the aggregated air time of leg $i$ since last ground contact \\
  \hline

  $ 1.0 \text{ or } 0.0 $  & * New reward for execution of a symmetric tripod gait; gives fixed reward if the correct set of feet are in ground contact \\
  \hline  

  \end{tabularx}

  \vspace{-5pt}

\end{table}

The network is included in a ROS 2 node to use the trained policy on the robot hardware and in a MuJoCo simulation.
The needed observations are collected from robot sensor readings or substituted with approximations based on state estimation.
The policy inference runs at 200~Hz and uses the following observations: linear acceleration and angular velocity of the robot base, the approximate gravity vector, commanded linear (x, y) and angular (around z) velocities, current joint positions and velocities as well as the previous inference results.
A list of reward function terms used in training is provided in Table~\ref{table:costs}, most of which originate in legged\_gym~\cite{rudin_legged_gym_2022}.
Most reward weights were optimized to improve locomotion for the novel robot.

To tackle the sim2real gap encountered when transferring trained policies onto the hardware, we are utilizing an Actuator Network as described by Hwangbo  et~al.\ \cite{hwangbo_learning_2019}.
For this, we collected actuator data on a per-joint basis while walking with the kinematic-based controller and trained an actuator network.
Additionally, we also captured non-walking motions designed to create a more varied data set of motor behaviors.
The actuator network is used in policy training to have the simulated actuators better match the hardware's behavior.
Some remaining minor oscillations are removed by applying a low-pass filter to the joint commands.

\section{RESULTS} \label{sec:results}

LAURON VI should be able to employ different control strategies to facilitate omnidirectional movement.
Even without external sensing capabilities, the robot should be able to traverse simple terrains outside a lab environment.
Additionally, it should be able to carry additional payloads of considerate weight, without the need of modeling the additional mass.

The robot is able to move through a lab environment using different controllers, showcasing its flexibility.
Additionally, the robot should be able to walk in a Mars analog environment to demonstrate its robustness and general feasibility for field missions.

\subsection{Laboratory testing}

The three presented controllers were used to walk forwards, backwards, sideways, and rotate around the robot's center.
The robot traversed flat terrain, and all controllers were able to compensate for small external impulses while standing and walking.
The maximum velocities of the controllers are presented in Table~\ref{table:velo}.

\begin{table}[thb]
  \renewcommand{\arraystretch}{1.3}
  \vspace{0.2cm}
  \caption{Maximum measured velocities for the kinematic-based, model-predictive control, and reinforcement learned strategies.}
  \label{table:velo}
  \centering
  \begin{tabular}{|c | c c c|}
  \hline
   & \multicolumn{3}{c|}{Maximum velocity} \\
  \thead{Controller} & \thead{forwards} & \thead{backwards} & \thead{sideways} \\
  \hline
  Kinematic-based & 0.17 $m/s$ & 0.16 $m/s$ & 0.05 $m/s$ \\
  \hline
  Model-predictive & 0.10 $m/s$ & 0.05 $m/s$ &  0.05 $m/s$ \\
  \hline
  Reinforcement Learned & 0.43 $m/s$ & 0.32 $m/s$ & 0.31 $m/s$ \\
  \hline
  \end{tabular}

  \vspace{-5pt}

\end{table}

The robot is able to utilize either two or four 22,2 V batteries with 355.2 Wh each.
Each battery consists of six cells with a nominal voltage of 3.7 V.
The robot achieved 108 minutes of continuous walking with two batteries before the first battery cell reached a 3.5 V threshold.
For this test, the robot was walking a course of 5 m forwards and backwards without turning.
The kinematic-based controller was used for this experiment.

\begin{figure}[b]
    \includegraphics[width=\columnwidth]{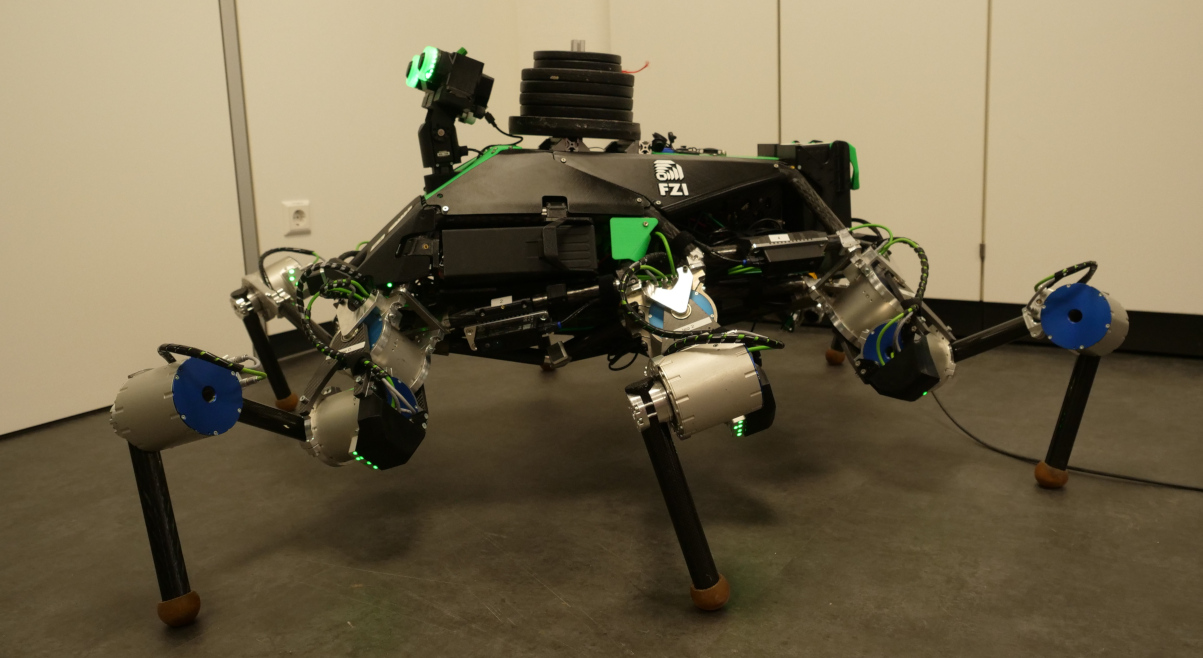}
    \caption{LAURON VI carrying an additional weight of 12 kg.}
    \label{fig:weight}
    \vspace{0.6cm}
\end{figure}

To show the robot's adaptability to carry different payloads and resist external influences, the MPC was tested while carrying or pulling additional weights.
The additional mass was not added to the robot model to introduce a model inaccuracy.
The MPC achieved stable walking in a tripod gait while 12 kg were added on the back of the robot, as shown in Fig.~\ref{fig:weight}.
This mass is considerably higher than a sensor payload.
For comparison, our navigation payload consisting of a lidar scanner and additional compute weights around 1.6 kg.
The MPC was also able to drag a box filled with 22~kg.

The experiments with the model-predictive controller have shown some inaccuracies regarding the modeling of the hardware that need to be addressed.
The control approach presented in Section~\ref{sec:MPC} is quite sensitive to changes to the center of mass.
To counteract this, a PD-Compensator is adopted, following the work of Chang, Ma, and Al~\cite{chang_quadruped_2021}, to increase the controller's robustness to model inaccuracies and external influences.

\subsection{Qualitative Evaluation in Mars analog mission}

As part of the publicly funded project IntelliRISK2, we evaluated LAURON VI in a Mars analog mission in Spain.
The Tabernas Desert in Spain was chosen as the test location, see Fig.~\ref{fig:Spain}.
The area used for the field test provided a diverse set of terrains, including flat areas, steep inclines, saline encrustations, as well as small and bigger stones and other obstacles.
For a more in-depth discussion of the encountered terrain and overall mission, see Puck et~al.\ for details \cite{puck_risk_2023}.

We used this mission to test the robot's design in this challenging environment.
Using the kinematic-based controller, the robot traversed different terrain types.
Given the robot's current lack of external sensing, the step and body heights were adapted to different ground types.
Besides that, no major configuration changes were needed to move from a lab environment to a real-world test.
Some weak points in the construction could be identified.
Most critical was the twisting of some carbon fiber tubes in the main body and one leg.
To address this issue, most connections were either glued or screwed securely after the mission to prevent further failures.

Besides showcasing the ability to traverse varied terrains, we could also verify the utility of hot-swappable batteries and the sensor-carrying capabilities of the robot in the field.

\section{CONCLUSIONS AND FUTURE WORKS} \label{sec:conclusion}

We've presented LAURON VI, a six-legged robot with 18 series elastic actuators designed for the research of dynamic six-legged locomotion.
Its leg configurations are designed to increase the individual and combined leg workspaces.
The main body's structure is topology-optimized to ensure stability even under heavy loads while reducing its weight to increase runtime.

The presented robot is flexible and powerful, allowing research into diverse locomotion strategies for six-legged systems,
as showcased by the successful utilization of kinematic-based, model-predictive, and reinforcement learned walking controllers.
The development of this insectoid robot with series elastic actuators focused on a lightweight construction,
aiming to provide a robust and future-oriented research platform for the development of dynamic gaits.
We showcased multiple control strategies on a single platform for cross-evaluation.
The robot is able to carry heavy sensors and mission payloads for long mission durations.
We aim to utilize LAURON VI in future projects as a research platform for autonomous operations as one part of our heterogeneous robot~team.

The development of LAURON VI has laid the foundation for future research into dynamic gaits and the transfer of the impressive results in quadrupeds to the hexapod domain.
We believe that the increased velocity and reactiveness will further the viability of hexapods as research platforms and their use in industrial applications.

Future improvements to the robot include the addition of proprioceptive and external sensors for improved state and environment estimation, which will increase the MPC performance.
By adding six depth cameras to capture its immediate surroundings, the locomotion in rough terrain will be improved.
The reinforcement learning will focus on synthesizing novel gait patterns for six-legged robots, facilitating the flexible kinematic structure and dynamic options.

\addtolength{\textheight}{-2.5cm}

\newpage

\bibliographystyle{template/IEEEtran} %
\bibliography{template/IEEEabrv,paper_lauron_vi}
\end{document}